\documentclass{article}

\usepackage[preprint]{neurips_2021}

\usepackage[utf8]{inputenc} 
\usepackage[T1]{fontenc}    
\usepackage{hyperref}       
\usepackage{url}            
\usepackage{booktabs}       
\usepackage{amsfonts}       
\usepackage{nicefrac}       
\usepackage{microtype}      
\usepackage{xcolor}         
\usepackage{graphicx}
\graphicspath{ {./images/} }
\usepackage{algorithm}
\usepackage[noend]{algpseudocode}
\title{Person Re-Identification with a Locally Aware Transformer}

\author{%
  Charu ~Sharma*, Siddhant R. ~Kapil*, David ~Chapman* \\
  {*} Contributed Equally \\
  Department of Computer Science\\
  University of Maryland Baltimore County\\
  Baltimore MD 21250 \\
  \{charus2, skapil1, dchapm2\}@umbc.edu\\

}

\begin{document}

\maketitle

\begin{abstract}

 Person Re-Identification is an important problem in computer vision-based surveillance applications, in which the same person is attempted to be identified from surveillance photographs in a variety of nearby zones.  At present, the majority of Person re-ID techniques are based on Convolutional Neural Networks (CNNs), but Vision Transformers are beginning to displace pure CNNs for a variety of object recognition tasks. The primary output of a vision transformer is a global classification token, but vision transformers also yield local tokens which contain additional information about local regions of the image. Techniques to make use of these local tokens to improve classification accuracy are an active area of research. We propose a novel Locally Aware Transformer (LA-Transformer) that employs a Parts-based Convolution Baseline (PCB)-inspired strategy for aggregating globally enhanced local classification tokens into an ensemble of $\sqrt{N}$ classifiers, where $N$ is the number of patches. An additional novelty is that we incorporate blockwise fine-tuning which further improves re-ID accuracy. LA-Transformer with blockwise fine-tuning achieves rank-1 accuracy of $98.27 \%$ with standard deviation of $0.13$ on the Market-1501 and $98.7\%$ with standard deviation of $0.2$ on the CUHK03 dataset respectively, outperforming all other state-of-the-art published methods at the time of writing.
\end{abstract}

\section{Introduction}
In recent years, Person Re-Identification(re-ID) has gained a lot of attention due to its foundational role in computer vision based video surveillance applications. Person re-ID is predominantly considered as a feature embedding problem. Given a query image and a large set of gallery images, person re-ID generates the feature embedding of each image and then ranks the similarity between query and gallery image vectors. This can be used to re-identify the person in photographs obtained by nearby surveillance cameras.

Recently, Vision Transformer (ViT) as introduced by \citet{ViT} is gaining substantial traction for image recognition problems. While some methods for image classification \citep{ViT, Deit}, and for image retrieval \citep{VitImageRetrieval} are focused only on the \textit{classification token}, some approaches utilize the fact that \textit{local tokens}, which are also outputs of the transformer encoder, can be used to improve performance of many computer vision applications including image segmentation \citep{visualTransformer, pyramidVT, transunet}, object detection \citep{ transformerbased_obj_detection, pyramidVT} and even person re-ID \citep{TransReID}. Nevertheless, at present, approaches to make use of \textit{local and global tokens} are an active area of research.

In the words of \citet{transformerbased_obj_detection}, \textit{"The remaining tokens in the sequence are used only as features for the final class token to attend to. However, \underline{these unused outputs correspond to the input patches}, and in theory, could encode local information useful for performing object detection"}.\citet{transformerbased_obj_detection} observed that the \textit{local tokens}, although theoretically influenced by global information, also have substantial correspondence to the original input patches. One might therefore consider the possibility of using these \textit{local tokens} as an enhanced feature representation of the original image patches to more strongly couple vision transformer encoders to fully connected (FC) classification techniques. This coupling of \textit{local patches} with FC classification techniques is the primary intuition behind the LA-Transformer architectural design.

Part-based Convolutional Baseline (PCB) \citep{PCB} is a strong convolutional baseline technique for person re-ID and has inspired many state-of-the-art models \citep{STReid, beyond_human_parts, pyramid}. PCB partitions the feature vector received from the backbone network into six vertical regions and constructs an ensemble of regional classifiers with a voting strategy to determine the predicted class label. A limitation of PCB is that each regional classifier ignores the global information which is also very important for recognition and identification. Nevertheless, PCB has achieved much success despite this limitation, and as such the design of LA-Transformer uses a PCB-like strategy to combine globally enhanced \textit{local tokens}. 

Our work also improves on the recent results of \citet{TransReID}, who was the first to employ Vision Transformers to person re-ID and achieved results comparable to the current state-of-the-art CNN based models. Our approach extends \citet{TransReID} in several ways but primarily because we aggregate the globally enhanced \textit{local tokens} using a PCB-like strategy that takes advantage of the spatial locality of these tokens. Although \citet{TransReID} makes use of \textit{fine-grained local tokens}, it does so with a ShuffleNet \citep{shufflenet} like Jigsaw shuffling step which does not take advantage of the 2D spatial locality information inherent in the ordering of the local tokens. LA-Transformer overcomes this limitation by using a PCB-like strategy to combine the globally enhanced \textit{local tokens} while first preserving their ordering in correspondence with the image dimension.

An additional novelty of our approach is the use of blockwise fine-tuning which we find is able to further improve the classification accuracy of LA-Transformer for person re-ID.  Blockwise fine-tuning is viable as a form of regularization when training models with a large number of parameters over relatively small in-domain datasets. \citet{ULMFit} advocate for blockwise fine-tuning or \textit{gradual unfreezing} particularly when training language models due to a large number of fully connected layers. As vision transformers also have high connectivity, we find that this approach is able to further improve the classification accuracy for LA-Transformer.

This paper is organized as follows: Firstly, we discuss related work involving Transformer architectures and other related methodologies in person re-ID. Secondly, we describe the architecture of LA-Transformer, including the novel locally aware network and blockwise fine-tuning techniques. Finally, we present quantitative results of the person re-ID including mAP and rank-1 analysis on the market-1501 and CUHK03 datasets.

\section{Related Work}

For many years CNN based models have dominated image recognition tasks including person re-ID.  A vast body of research has been performed to determine the best strategy to extract features using CNNs to address issues like appearance ambiguity, background perturbance, partial occlusion, body misalignment, viewpoint changes, and pose variations, etc.  \citet{posesensitive} proposed a Pose-Sensitive Embedding to incorporate information associated with poses of a person in the model, \citet{conditionalEmbedding} used a Graph Convolution Network \citep{GCN} to generate a conditional feature vector based on the local correlation between image pairs, \citet{lightweight} used global channel-based and part-based features, \citet{alignedreid} used global pooling to extract global features and horizontal pooling followed by $1 \times 1$ CNN for local features.  CNN based methods have led to many advances in recent years and are continuing to be developed for person re-ID.

Another branch of techniques for person re-ID focuses on the development of highly engineered network designs that incorporate additional domain knowledge to improve re-ID performance.  \citet{PartAware} used a part-aware approach for which the model performs the main task as well as auxiliary tasks for each body part. \citet{ViewpointVehicleReID} and \citet{Viewpoint} use viewing angles as additional features. \citet{PartLoss}  introduced the idea of calculating part loss and \citet{PCB} (Part-based Convolutional Backbone a.k.a. PCB) improved on it. Even current top-performing models like \citet{STReid} used PCB along with domain-specific Spatio-temporal distribution information to achieve good results on the Market-1501 dataset. In our work we incorporate PCB-like local classifiers with Vision Transformers, and furthermore we find that our model performs better if we pass global information along with local features.  LA-Transformer achieves results with comparable and slightly higher rank-1 accuracy than the reported results of \citet{STReid} over Market-1501 and does so without the use of additional Spatio-temporal information.

Interest in Vision Transformers grew initially from attention mechanisms which were first employed for language translation problems in NLP \citep{firstAttention}, Attention mechanisms have been employed to great effect in image recognition. \citet{spatialattention} introduced parameter-free spatial attention to integrating spatial relations to Global Average Pooling (GAP). \citet{SAM_CAM} used Spatial Attention Module (SAM), and Channel Attention Module (CAM) to deliver prominent spatial and channel information. \citet{abdnet} propose Position Attention Module (PAM) for semantically related pixels in the spatial domain along with CAM.  Attention mechanisms continue to be an active area of research for many problems related to object detection and recognition.

Transformers were first introduced in NLP problems by \citet{Transformer}, and now Transformers are contributing to many new developments in machine learning. \citet{ViT} introduced transformers to images by treating a 16x16 patch as a word and treating image classification as analogous to text classification. This approach showed promising results on ImageNet and it was soon adopted in many image classification problems \citep{image_tranformer, nonlocal}.  Object detection is another highly related problem for which vision transformers have been recently applied  \citep{DETR, transformerbased_obj_detection}.   \citet{transformerbased_obj_detection} described a correspondence between local tokens and input patches and combined local tokens to create spatial feature maps. At present, this observation of the correspondence between \textit{local tokens} and input patches has yet to be applied to a wide variety of computer vision problems, nor has it been previously explored in the context of person re-ID.  One exception is in the area of image segmentation, for which recent works are beginning to take advantage of the 2D ordering of the local tokens in order to produce more accurate predicted masks \citep{visualTransformer, pyramidVT, transunet}. Our approach builds upon the recent work of \citet{TransReID} who was the first to apply vision transformers to object and person re-ID.  Although the approach of \citet{TransReID} makes use of global and local tokens, \citet{TransReID} combines the local tokens using a jigsaw classification branch which shuffles the ordering of the local features.  Shuffling the order of local features does not take advantage of the observation of \citet{transformerbased_obj_detection} in that local features correspond strongly with input patches and therefore have a natural ordering in the form of a 2D image grid.  Conversely, LA-Transformer takes advantage of the spatial locality of these local features by combining globally enhanced \textit{local tokens} with a PCB-like strategy \citep{PCB}.  Furthermore, LA-Transformer incorporates the blockwise fine-tuning strategy as described by \cite{ULMFit} as a form of regularization for high-connectivity pre-trained language models.  As such LA-Transformer builds upon recent advances in the application of vision transformers in tandem with novel training techniques to achieve state of the art accuracy in person re-ID.

\section{Methodology}

LA-Transformer combines vision transformers with an ensemble of FC classifiers that take advantage of the 2D spatial locality of the globally enhanced local tokens.  Section \ref{LA Transformer} describes the overall architecture including the backbone vision transformer (section \ref{part1}), as well as the PCB inspired classifier network ensemble (section \ref{part2}).  The blockwise fine-tuning strategy is described in section \ref{Blockwise}.  As such, these sections describe the major elements of the LA-Transformer methodology.
\label{headings}

\begin{figure}[htp]
    \includegraphics[width=14cm,height=10cm]{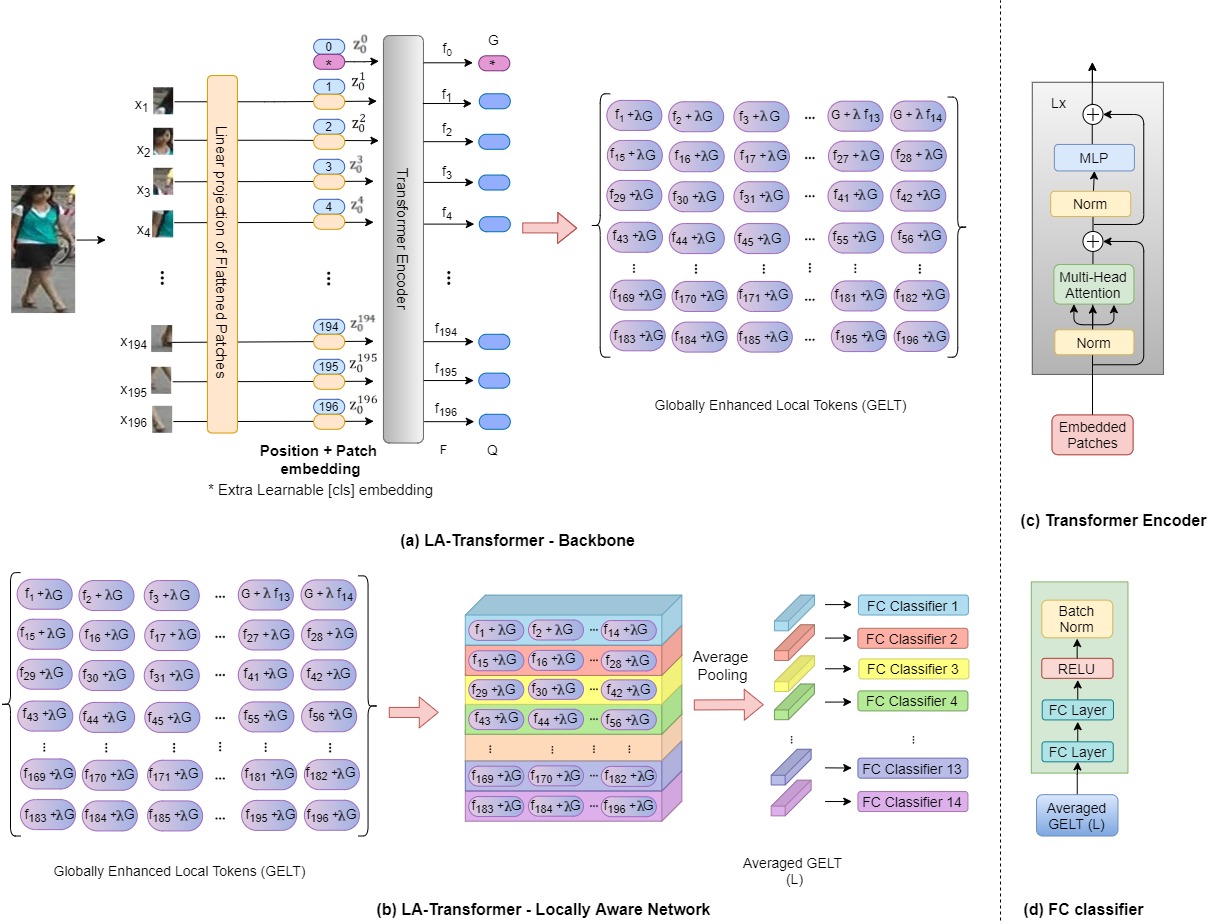}
    \label{fig1:Model}
    \caption{Architecture of LA-Transformer. Part \textbf{(a)} - shows the backbone architecture. The input image is converted into patch embedding using 2D convolution. The class embedding (cls embedding) is prepended to the patch embedding. Then the position embeddings are added and this resulting sequence is fed to the transformer encoder. $F = {f_0, f_1,...,f_N}$ is the output of the transformer encoder where $f_0$ is the global vector $G$ and remaining tokens from $f_1$ to $f_{N}$ are local tokens $Q$. $G$ and $Q$ are then combined using weighted averaging and are called Globally Enhanced Local Tokens (GELT). GELT are then arranged into a 2D spatial grid. Part \textbf{(b)} - shows the Locally Aware Network. $L$ is the row-wise average of GELTs and is performed using average pooling. $L$ is then fed to the locally aware classification ensemble. Part \textbf{(c)} and \textbf{(d)} describes the architecture of transformer encoder and FC classifier respectively.
    }
\end{figure}

\subsection{Locally Aware Transformer}\label{LA Transformer}
LA-Transformer (figure \ref{fig1:Model}) consists of two main parts: a backbone network and a locally aware network. Both components are interconnected and trained as a single neural network model.  

The backbone network is the ViT architecture as proposed by \citet{ViT}.
ViT generates tokens $F = {f_0, f_1,.., f_N}$. The token $f_0$, also known as the global classification token and we refer to this token as the \textit{global token} $G$.  Supplementary outputs ${f_1..f_{196}}$ are referred to as \textit{local tokens} which we denote to collectively as $Q$. Globally Enhanced Local Tokens (GELT) are obtained by combining \textit{global tokens} and \textit{local tokens} ($G$ and $Q$) using weighted averaging and are arranged into a $14\times 14$ 2D spatial grid as seen in Figure \ref{fig1:Model}(a). The row-wise averaged GELTs are then fed to the locally aware classification ensemble as seen in Figure \ref{fig1:Model}(b) to classify during the training process and to generate feature embedding (by concatenating L) during the testing process. These steps are described in greater detail in the following sections \ref{part1} and \ref{part2}

\subsubsection{LA-Transformer Backbone} \label{part1}
The backbone network of LA-Transformer is the ViT vision transformer \citep{ViT}.  ViT requires extensive training data on the order of $14M - 300M$ images to train effectively, but the Market1501 and CUHK-03 datasets are relatively small (Table~\ref{DatasetOverview}) in comparison on the order of $10$'s of thousands of images. As such we employed a pre-trained ViT model, and further made use of blockwise fine-tuning to improve accuracy as described in section \ref{Blockwise}

\paragraph{Embeddings} The backbone ViT architecture takes images of size $224\times 224$ as input, and as such the Market1501 and CUHK-03 images are re-sampled to this resolution during training. First, the image is converted into $N$ number of patches $x^i_{p}|i=1,..,N$. Each patch is then linearly projected into $D$ dimensions using the \textit{patch embedding} function ($E(x^i_{p})|i=1,..,N$) (eq.~\ref{eqn:13}), which is obtained using a convolution layer with a kernel size of $16\times 16$.  For non-overlapping patches, a stride equal to $16$ is used. $D$ is the number of channels and is set to $768$ which represents the size of the embedding. The total number of patches $N$ depends on kernel size, stride, padding, and size of the image.  $N$ can be easily calculated using the eq.~\ref{eqn:3}. Assuming padding is 0, and ${H}$, ${W}$ are height and width of an image, ${K_H}$, ${K_W}$ are height and width of the kernel and $S$ is kernel stride. 

\begin{equation}
\label{eqn:3}
N = \left(\frac{H-K_H}{S} +1\right)\times\left(\frac{W-K_W}{S} + 1\right)
\end{equation}

Afterward, the learnable \textit{class embedding} $x_{class}$ is prepended with the patch embedding ($E(x^i_{p}$)) whose output state keeps the information of the entire image and serves as the global vector. The resulting vectors are then added with \textit{position embeddings} $P$ to preserve the positional information. Subsequently, the final sequence of vectors $z_0$ (eq.~\ref{eqn:13}) is fed into the transformer encoder (figure\ref{fig1:Model}) to generate $N+1$ feature vectors where $N$ is the number of patches plus class embedding.

\begin{equation}
\label{eqn:13}
z_0 = [x_{class}; E(x^1_{p}); E(x^2_{p}); .... ;E(x^N_{p})] + P
\end{equation}

\paragraph{Transformer Encoder}  The transformer encoder consist of total $B=12$ blocks. Each block contains alternating MSA (Multiheaded Self-Attention) introduced by \citet{Transformer} and MLP blocks. The Layernorm (LN) is applied before MSA and MLP blocks and a residual connections is applied after each encoder block. The output of transformer encoder $F$ described in eq.~\ref{eqn:11} passes through all the $B$ blocks (eq.~\ref{eqn:10} and \ref{eqn:12} ). 

\begin{equation}
\label{eqn:10}
z'_b = z_{b-1} + MSA(LN(z_{b-1})) 
\end{equation}
\begin{equation}
\label{eqn:12}
z_b = z'_{b} + MLP(LN(z'_{b})) 
\end{equation}
\begin{equation}
\label{eqn:11}
F = LN(z_B)
\end{equation}

While the seminal work of \citet{ViT} only uses classification token $z^0_B$ for classification, LA-Transformer makes use of all of the features $z_B$ eq.~\ref{eqn:11}. Though the class embedding can be removed from the backbone network, our experiments show promising results with class embedding serving as a global vector (Table~\ref{ModelsOverview}). From our experiments, it is clear that ViT as a backbone network is a good choice for person re-ID based problems. Further, we believe that any transformer based model like Diet by \citet{Deit}, or DeepViT by \citet{DeepViT} can be used as a backbone network.

\subsubsection{Locally Aware Network}\label{part2}

The Locally Aware Network is a classifier ensemble similar to the PCB technique of \citet{PCB} but with some differences. Firstly, in \citet{PCB} the input features are purely local, whereas in LA-Transformer, we find that the inclusion of global vectors along with local vectors via weighted averaging can increase the network accuracy.  Secondly, although in \citet{PCB} the image is divided into six input regions, we divide the 2D spatial grid of tokens into $\sqrt{N} = 14$ regions as seen in Figure \ref{fig1:Model}.  Finally, while PCB uses a convolutional backbone, LA-Transformer uses the ViT backbone.

In Figure \ref{fig1:Model}, the transformer encoder outputs $N+1$ feature vectors. The global tokens $G = f_0$ and local tokens $Q = [f_1, f_2, f_3,...,f_N]$ are obtained for which $N$ is number of patches. $N_R$ is defined as the total number of patches per row and $N_C$ as the total number of patches per column. In our case, $N_R = N_C = \sqrt{N}$. Then we define $L$ as the averaged GELT obtained after average pooling of $Q$ and $G$ as follows,

\begin{equation}
\label{eqn:1}
L_i = \frac{1}{N_R} \sum_{j=i * N_R +1}^{(i+1) * N_R } \frac{(Q_j +\lambda G)}{(1+\lambda)} \qquad \qquad   i=0...N_C-1
 \end{equation}

In eq.~\ref{eqn:1} all the patches in a row are averaged to create one local vector per row . The total number of FC classifiers is equal to $N_C$. Each FC classifier contains two fully connected layers with RELU and Batch Normalization. We define $y$ as the output of LA-Transformer as follows,

\begin{equation}
\label{eqn:2}
y_i = FC_i( L_i) \qquad \qquad i=1...N_C
\end{equation}

The outputs $y$ are passed through softmax and the softmax scores are summed together. The argument of the maximum score represents the ID of the person as follows.

\begin{equation}
score = \sum_{i=0}^{N_C} softmax(y_i)
\end{equation}
\begin{equation}
prediction = argmax(score)
\end{equation}

\subsection{Fine-tuning Strategies}
According to the recent studies of \citet{Deit} and \citet{ViT}, training a vision transformer from scratch requires about 14M-300M images. Person re-ID datasets are known for their small size and training a transformer on these datasets can quickly lead to overfitting.  As such, ViT was pre-trained on ImageNet (\citet{imagenet21k}), and then fine-tuned on person re-ID datasets. Blockwise fine-tuning was applied which is highly similar to the gradual unfreezing method described by \citet{ULMFit} for the purposes of training large language models in the event of limited training data from a target domain. 

\paragraph{Blockwise Fine-tuning}\label{Blockwise}
In blockwise fine-tuning, all transformer blocks are frozen in the start except for the bottleneck model. After every $t$ epochs (where $t$ is a hyper-parameter), one additional transformer encoder block is unfrozen and the learning rate is reduced as described by Alg\ref{algo}. Blockwise fine-tuning helps in mitigating the risk of catastrophic forgetting of the pre-trained weights \citep{ULMFit}. The learning rate decay helps in reducing the gradient flow in the subsequent layers hence prevent abrupt weight updates. 

\begin{algorithm}
\label{algo}
\caption{Blockwise Fine-tuning}
\begin{algorithmic}[1]
\State Freeze all the transformer blocks $B$
\State \textbf{Initialize parameters:} $t = 2$, $b = 12$, $lr=3e-4$, $lr-decay = 0.85$
\While{$ 0 <= i < epochs$}  
    \If{$i\%t==0 \ and \ b >0 $}  \Comment{ViT has 12 blocks}
        \State $unfreeze \ B[b]$  \Comment Unfreeze the last block first
        \State $b \leftarrow b-1$  \Comment Reduce block index counter
        \State $lr \leftarrow lr*lr\_decay$ \Comment Reduce lr rate
    \EndIf
\EndWhile 
\end{algorithmic}
\end{algorithm}

\section{Experiments}

\subsection{Datasets and Metrics}

\begin{table} 
  \label{DatasetOverview}
    \caption{Datasets Overview}
  \centering
  \begin{tabular}{clclclc}
    \toprule
    Dataset     & Classes     & Train     & Query     & Gallery \\
    \midrule
    Market-1501 & 751       & 12192      & 3368      & 19744 \\
    CUHK-03     & 1367      & 13131     & 965       & 965  \\
    \bottomrule
  \end{tabular}
\end{table}

\paragraph{Datasets} LA Transformer is trained over two benchmark datasets; Market-1501 and CUHK-03. Table~\ref{DatasetOverview} gives the overview of datasets used to train the model. The Market-1501 dataset \citep{Market1501} contains total 1501 classes/identities captured by six different cameras.  Out of 1501 classes, the train set contains 750 classes, and the test set consists of 751 classes. A total of 12,192 images are present in the train set. The test set is divided into a query set of 3,368 images and a gallery set of 19744 images. CUHK-03 dataset \citep{cuhk03} contains a total of 1,367 classes captured by six cameras. There are 13,131 images in the train set and 1,930 images in the test set (965 in query and 965 in the gallery set).

\paragraph{Evaluation protocol} By convention, re-ID is evaluated over two standard evaluation metrics; Cumulative Matching Characteristics (CMC) and Mean Average Precision (mAP). We apply these metrics to assess the performance of the LA-Transformer and other experiments. 

\subsection{Model Implementation Details} \label{ModelImplementation}

ViT was pre-trained on ImageNet-21K and used as a backbone network as well as a baseline model \citep{ViT, imagenet21k}. All the images are resized into $224 \times 224$ as this resolution is compatible with the backbone network.  The model is trained over 30 epochs with a batch size of 32. We used the Adam optimizer with an initial learning rate of $3e-5$, step decay of $0.8$, ${t} = 2$  and ${\lambda} = 0.8$. For testing, we concatenated all of the averaged GELTs $L$ to generate the feature embedding. To efficiently calculate the Euclidean norm between the query and gallery vectors, we use the FAISS library \citet{FAISS}. All the models are trained and tested on a single GPU machine with an Nvidia RTX2080 Ti with 11 GB VRAM, and 64 GB RAM.

\begin{table}
  \caption{Ablation result of the influence of global and local features on baseline ViT and LA-Transformer with and without blockwise fine-tuning over Market-1501}
  \centering
  \begin{tabular}{clclclclclclcl}
    \toprule
    \multicolumn{4 }{c}{} & \multicolumn{2}{c}{Without BW-FT} & \multicolumn{2}{c}{With BW-FT} \\ 
\cmidrule(r){5-6} \cmidrule(r){7-8}
    \# & Model  & Classifiers     & Tokens   & Rank-1     & mAP     & Rank-1     & mAP \\
    \midrule
    1 & ViT  & 1    & Global  & 95.8	& 89.55 & 96.2 & 90.5    \\
    2 & ViT     & 1  & Local & 95.1 & 86.5	& 95.45	& 87.8 \\
    3 & ViT     & 1  & Global+Local & 96.3	& 90.1	& 96.6	& 90.30  \\
    4 & LA & 14 & Global   & 96.9	& 92.5	& 97.74	& 92.68      \\
    5 & LA & 14 & Local    &  96.1	& 91.1	& 97.23	& 91.93   \\
  6 & LA & 14 & Global+ Local  & 97.55	& 93.3	& 98.27	&  94.46   \\
    \bottomrule
  \end{tabular}
  \label{ModelsOverview}
\end{table}

\begin{figure}[htp]
    \centering
    \includegraphics[width=14cm,height=5cm]{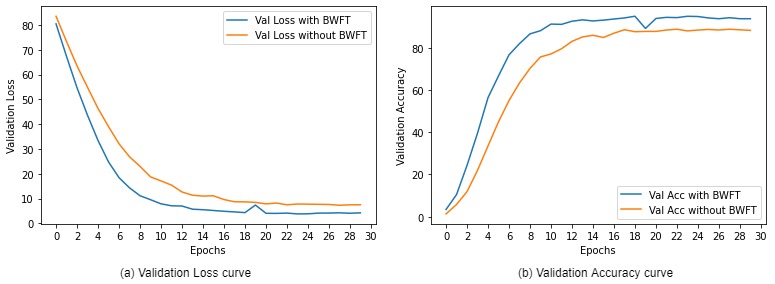}
    \label{fig:graph}
    \caption{Part(\textbf{a})- compares the validation loss of LA-Transformer with and without Blockwise Fine-tuning (BWFT). Part(\textbf{b}) - compares validation accuracy of LA-Transformer with and without BWFT.  Both graphs show results using Market-1501. }
\end{figure}

\subsection{Ablation study of LA Transformer}
The table~\ref{ModelsOverview} compares the performance of variations of LA-Transformer versus the same variations of baseline ViT using the Market-1501 dataset. All six experiments are performed with and without blockwise fine-tuning. \textit{Experiment 1} is the baseline model that uses only the global token to generate feature embedding. \textit{Experiment 2} uses only the local tokens of the transformer encoder and exhibits the lowest rank-1 accuracy ($95.1$) and mAP score ($86.5$) out of all of the variations of ViT.  \textit{Experiment 3} combines the the first and second experiments by utilizing the globally enhanced local tokens. The impact of global and local features is also compared using LA-Transformer via three variations:  global, local, and globally enhanced local tokens.  All of the experiments with LA-Transformer perform better than the baseline ViT and its variations. LA-Transformer increases the rank-1 accuracy by ${+2.3\%}$  and mAP score by ${+2.6}$ on an average versus ViT over the experiments in table~\ref{ModelsOverview}. Similar to ViT with local features, LA-Transformer with only local features achieves the lowest accuracy of the LA-Transformer only experiments.  Therefore, we conjecture that using only local vectors to predict the output and generate the final embedding is not sufficient. Nevertheless, using globally enhanced local tokens outperforms the local only results by ${+1.31\%}$ rank-1 and ${+0.265}$ mAP and improves over the global only results by ${+0.62\%}$ rank-1 and ${+0.5}$ mAP. Therefore LA-Transformer using globally enhanced local tokens achieves the highest rank-1 and mAP scores of all technique $+$ feature embedding designs in this comparison.

\paragraph{Ablation study of Blockwise fine-tuning}
Blockwise fine-tuning achieves higher rank-1 and mAP scores in all experiments as compared against similar experiments without blockwise fine-tuning over the Market-1501 dataset. As seen in Table~\ref{ModelsOverview}, blockwise fine-tuning increases the rank-1 accuracy by ${+0.6\%}$ and mAP score by  ${+0.77}$ on average across all of the experiments in this Ablation study. During blockwise fine-tuning, the hyperparameter $t$ is set to 2 which means, after every 2 epochs one additional block is unfrozen. The baseline ViT model has 12 blocks. Therefore, it takes 22 epochs to unfreeze and train on all the layers. However, for most models, we found the best validation score is reached before the 22nd epoch, but rather after the 18th epoch yielding 10 trainable blocks during fine tuning. Figure \ref{fig:graph} shows the comparison of validation results for LA-Transformer trained with and without blockwise fine-tuning. It can be clearly seen that blockwise fine-tuning leads to faster convergence and better results than the training model without blockwise fine-tuning.

\begin{table}
  \caption{Comparison with State of the Art methods}
  
  \centering
  \begin{tabular}{clclclc}
    \toprule
    \multicolumn{3}{c}{Market-1501} & \multicolumn{3}{c}{CUHK-03} \\ 
\cmidrule(r){1-3} \cmidrule(r){4-6}
    Model  & Rank-1 & mAP & Model & Rank-1 & mAP\\
    \midrule
    PCB	& 92.3	& 77.4 &UTAL  & 56.3 & 42.3\\
    AANet &	93.9	& 83.4 & k-reciprocal 46  & 61.6 & 67.6\\
    IANet	& 94.4 &	83.1 & DG-Net 61.1 65.6 \\
    DG-Net &	94.8	& 86 & OIM & 77.5 & 72.5 \\
    TransReID & 95.2 & 88.9 & VI+LSRO & 84.6 & 87.6 \\
    Flip reid	& 95.8	& 94.7 & TriNet & 89.63 & - \\
    VA-reid	& 96.79	& 95.4  & FD-GAN & 92.6 & 91.3 \\
    st-Reid	& 98	& 95.5  & DCDS & 95.8 & - \\
    CTL Model	& 98 &	\textbf{98.3} & AlignedReID & 97.8 & - \\

    LA-Transformer  & \textbf{98.27}	&  94.46 & LA-Transformer  & \textbf{98.7} & \textbf{96.4} \\
    \bottomrule
  \end{tabular}
  \label{SOTA}
\end{table}

\subsection{Comparison with state of the art} \label{SOTAComparisonSection}
To evaluate the performance of LA-Transformer, it is trained and evaluated five times on Market-1501 and CUHK03 and the mean results are reported in Table~\ref{SOTA}. On Market-1501, the rank-1 accuracy of LA-Transformer is  $98.27\%$ with standard deviation of $0.13$ with blockwise fine-tuning and $97.55$ with standard deviation of $0.49$ without blockwise fine-tuning. On CUHK03, the rank-1 accuracy of LA-Transformer is $98.7\%$ with a standard deviation of $0.2$ with blockwise fine-tuning.

Table \ref{SOTA} compares LA Transformer with the state-of-the-art (SOTA) models on two benchmarks of person re-ID; Market-1501 and CUHK-03. On the Market-1501 dataset, LA-transformer achieves the highest reported rank-1 accuracy of all models in this comparison, and outperforms the rank-1 accuracy of the next highest SOTA model by $+0.27\%$.  On Market-1501, the mAP score lies among the top five SOTA models. In case of CUHK-03, LA-Transformer achieves both the highest rank-1 accuracy as well as the highest mAP score, and outperforms the next highest SOTA models score by $+0.9\%$ (rank-1) and $+5.1$ (mAP) respectively.

\section{Conclusion}
We present a novel technique for person re-ID called Locally Aware Transformer (LA-Transformer) which achieves state of the art performance on the Market-1501 and CHUK-03 datasets.  This approach makes two contributions toward solving the person re-ID problem. First, we show that the global token and local token outputs of vision transformers can be combined with a PCB-like strategy to improve re-ID accuracy.  Secondly, we incorporate blockwise fine-tuning to regularize the fine tuning of a pre-trained vision transformer backbone network.  We believe that vision transformers will continue to have a major positive impact in the field of computer vision, and we are hopeful that the architectural design of LA-transformer will lead to further innovation and the development of new and novel techniques to advance our understanding of person re-ID.

{
\small
\bibliographystyle{plainnat}
\interlinepenalty=10000
\bibliography{bibliography.bib}

\begin{thebibliography}{36}
\providecommand{\natexlab}[1]{#1}
\providecommand{\url}[1]{\texttt{#1}}
\expandafter\ifx\csname urlstyle\endcsname\relax
  \providecommand{\doi}[1]{doi: #1}\else
  \providecommand{\doi}{doi: \begingroup \urlstyle{rm}\Url}\fi

\bibitem[Bahdanau et~al.(2016)Bahdanau, Cho, and Bengio]{firstAttention}
Dzmitry Bahdanau, Kyunghyun Cho, and Yoshua Bengio.
\newblock Neural machine translation by jointly learning to align and
  translate, 2016.

\bibitem[Beal et~al.(2020)Beal, Kim, Tzeng, Park, Zhai, and
  Kislyuk]{transformerbased_obj_detection}
Josh Beal, Eric Kim, Eric Tzeng, Dong~Huk Park, Andrew Zhai, and Dmitry
  Kislyuk.
\newblock Toward transformer-based object detection, 2020.

\bibitem[Carion et~al.(2020)Carion, Massa, Synnaeve, Usunier, Kirillov, and
  Zagoruyko]{DETR}
Nicolas Carion, Francisco Massa, Gabriel Synnaeve, Nicolas Usunier, Alexander
  Kirillov, and Sergey Zagoruyko.
\newblock End-to-end object detection with transformers, 2020.

\bibitem[Chen et~al.(2021)Chen, Lu, Yu, Luo, Adeli, Wang, Lu, Yuille, and
  Zhou]{transunet}
Jieneng Chen, Yongyi Lu, Qihang Yu, Xiangde Luo, Ehsan Adeli, Yan Wang, Le~Lu,
  Alan~L. Yuille, and Yuyin Zhou.
\newblock Transunet: Transformers make strong encoders for medical image
  segmentation, 2021.

\bibitem[Chen et~al.(2019)Chen, Ding, Xie, Yuan, Chen, Yang, Ren, and
  Wang]{abdnet}
Tianlong Chen, Shaojin Ding, Jingyi Xie, Ye~Yuan, Wuyang Chen, Yang Yang, Zhou
  Ren, and Zhangyang Wang.
\newblock Abd-net: Attentive but diverse person re-identification, 2019.

\bibitem[Ding et~al.(2020)Ding, Wang, Wang, and Tao]{PartAware}
Changxing Ding, Kan Wang, Pengfei Wang, and Dacheng Tao.
\newblock Multi-task learning with coarse priors for robust part-aware person
  re-identification.
\newblock In \emph{Computer Vision and Pattern Recognition (cs.CV)}, 2020.
\newblock URL \url{arXiv:2003.08069}.

\bibitem[Dosovitskiy et~al.(2020)Dosovitskiy, Beyer, Kolesnikov, Weissenborn,
  Zhai, Unterthiner, Dehghani, Minderer, Heigold, Gelly, Uszkoreit, and
  Houlsby]{ViT}
Alexey Dosovitskiy, Lucas Beyer, Alexander Kolesnikov, Dirk Weissenborn,
  Xiaohua Zhai, Thomas Unterthiner, Mostafa Dehghani, Matthias Minderer, Georg
  Heigold, Sylvain Gelly, Jakob Uszkoreit, and Neil Houlsby.
\newblock An image is worth 16x16 words: Transformers for image recognition at
  scale.
\newblock In \emph{Computer Vision and Pattern Recognition (cs.CV)}, 2020.
\newblock URL \url{arXiv:2010.11929}.

\bibitem[El-Nouby et~al.(2021)El-Nouby, Neverova, Laptev, and
  Jégou]{VitImageRetrieval}
Alaaeldin El-Nouby, Natalia Neverova, Ivan Laptev, and Hervé Jégou.
\newblock Training vision transformers for image retrieval, 2021.

\bibitem[Guo et~al.(2019)Guo, Yuan, Huang, Zhang, Yao, and
  Han]{beyond_human_parts}
Jianyuan Guo, Yuhui Yuan, Lang Huang, Chao Zhang, Jinge Yao, and Kai Han.
\newblock Beyond human parts: Dual part-aligned representations for person
  re-identification, 2019.

\bibitem[He et~al.(2021)He, Luo, Wang, Wang, Li, and Jiang]{TransReID}
Shuting He, Hao Luo, Pichao Wang, Fan Wang, Hao Li, and Wei Jiang.
\newblock Transreid: Transformer-based object re-identification, 2021.

\bibitem[Herzog et~al.(2021)Herzog, Ji, Teepe, Hörmann, Gilg, and
  Rigoll]{lightweight}
Fabian Herzog, Xunbo Ji, Torben Teepe, Stefan Hörmann, Johannes Gilg, and
  Gerhard Rigoll.
\newblock Lightweight multi-branch network for person re-identification, 2021.

\bibitem[Howard and Ruder(2018)]{ULMFit}
Jeremy Howard and Sebastian Ruder.
\newblock Fine-tuned language models for text classification.
\newblock \emph{CoRR}, abs/1801.06146, 2018.
\newblock URL \url{http://arxiv.org/abs/1801.06146}.

\bibitem[Johnson et~al.(2017)Johnson, Douze, and J{\'e}gou]{FAISS}
Jeff Johnson, Matthijs Douze, and Herv{\'e} J{\'e}gou.
\newblock Billion-scale similarity search with gpus.
\newblock \emph{arXiv preprint arXiv:1702.08734}, 2017.

\bibitem[Kipf and Welling(2017)]{GCN}
Thomas~N. Kipf and Max Welling.
\newblock Semi-supervised classification with graph convolutional networks,
  2017.

\bibitem[Li et~al.(2014)Li, Zhao, Xiao, and Wang]{cuhk03}
Wei Li, Rui Zhao, Tong Xiao, and Xiaogang Wang.
\newblock Deepreid: Deep filter pairing neural network for person
  re-identification.
\newblock In \emph{CVPR}, 2014.

\bibitem[Parmar et~al.(2018)Parmar, Vaswani, Uszkoreit, Łukasz Kaiser,
  Shazeer, Ku, and Tran]{image_tranformer}
Niki Parmar, Ashish Vaswani, Jakob Uszkoreit, Łukasz Kaiser, Noam Shazeer,
  Alexander Ku, and Dustin Tran.
\newblock Image transformer, 2018.

\bibitem[Ridnik et~al.(2021)Ridnik, Ben-Baruch, Noy, and
  Zelnik-Manor]{imagenet21k}
Tal Ridnik, Emanuel Ben-Baruch, Asaf Noy, and Lihi Zelnik-Manor.
\newblock Imagenet-21k pretraining for the masses, 2021.

\bibitem[Sarfraz et~al.(2018)Sarfraz, Schumann, Eberle, and
  Stiefelhagen]{posesensitive}
M.~Saquib Sarfraz, Arne Schumann, Andreas Eberle, and Rainer Stiefelhagen.
\newblock A pose-sensitive embedding for person re-identification with expanded
  cross neighborhood re-ranking, 2018.

\bibitem[Sun et~al.(2018)Sun, Zheng, Yang, Tian, and Wang]{PCB}
Yifan Sun, Liang Zheng, Yi~Yang, Qi~Tian, and Shengjin Wang.
\newblock Beyond part models: Person retrieval with refined part pooling (and a
  strong convolutional baseline), 2018.

\bibitem[Touvron et~al.(2020)Touvron, Cord, Douze, Massa, Sablayrolles, and
  J{\'{e}}gou]{Deit}
Hugo Touvron, Matthieu Cord, Matthijs Douze, Francisco Massa, Alexandre
  Sablayrolles, and Herv{\'{e}} J{\'{e}}gou.
\newblock Training data-efficient image transformers {\&} distillation through
  attention.
\newblock \emph{CoRR}, abs/2012.12877, 2020.
\newblock URL \url{https://arxiv.org/abs/2012.12877}.

\bibitem[Vaswani et~al.(2020)Vaswani, Shazeer, Parmar, Uszkoreit, Jones, Gomez,
  Kaiser, and Polosukhin]{Transformer}
Ashish Vaswani, Noam Shazeer, Niki Parmar, Jakob Uszkoreit, Llion Jones,
  Aidan~N. Gomez, Lukasz Kaiser, and Illia Polosukhin.
\newblock Attention is all you need.
\newblock In \emph{Neural Information Processing Systems}, 2020.
\newblock URL \url{arXiv:2010.11929}.

\bibitem[Wang et~al.(2018{\natexlab{a}})Wang, Fan, Wang, Jiao, and
  Schiele]{spatialattention}
Haoran Wang, Yue Fan, Zexin Wang, Licheng Jiao, and Bernt Schiele.
\newblock Parameter-free spatial attention network for person
  re-identification, 2018{\natexlab{a}}.

\bibitem[Wang et~al.(2021)Wang, Xie, Li, Fan, Song, Liang, Lu, Luo, and
  Shao]{pyramidVT}
Wenhai Wang, Enze Xie, Xiang Li, Deng-Ping Fan, Kaitao Song, Ding Liang, Tong
  Lu, Ping Luo, and Ling Shao.
\newblock Pyramid vision transformer: A versatile backbone for dense prediction
  without convolutions, 2021.

\bibitem[Wang et~al.(2018{\natexlab{b}})Wang, Girshick, Gupta, and
  He]{nonlocal}
Xiaolong Wang, Ross Girshick, Abhinav Gupta, and Kaiming He.
\newblock Non-local neural networks, 2018{\natexlab{b}}.

\bibitem[Wu et~al.(2020)Wu, Xu, Dai, Wan, Zhang, Yan, Tomizuka, Gonzalez,
  Keutzer, and Vajda]{visualTransformer}
Bichen Wu, Chenfeng Xu, Xiaoliang Dai, Alvin Wan, Peizhao Zhang, Zhicheng Yan,
  Masayoshi Tomizuka, Joseph Gonzalez, Kurt Keutzer, and Peter Vajda.
\newblock Visual transformers: Token-based image representation and processing
  for computer vision, 2020.

\bibitem[Xie et~al.(2020)Xie, Wu, Zhang, Zhao, and Li]{SAM_CAM}
Ben Xie, Xiaofu Wu, Suofei Zhang, Shiliang Zhao, and Ming Li.
\newblock Learning diverse features with part-level resolution for person
  re-identification, 2020.

\bibitem[Yao et~al.(2017)Yao, Zhang, Zhang, Li, and Tian]{PartLoss}
Hantao Yao, Shiliang Zhang, Yongdong Zhang, Jintao Li, and Qi~Tian.
\newblock Deep representation learning with part loss for person
  re-identification.
\newblock In \emph{Computer Vision and Pattern Recognition (cs.CV)}, 2017.
\newblock URL \url{arXiv:1707.00798}.

\bibitem[Yao et~al.(2018)Yao, Zhang, Zhang, Li, and Tian]{STReid}
Hantao Yao, Shiliang Zhang, Yongdong Zhang, Jintao Li, and Qi~Tian.
\newblock Spatial-temporal person re-identification.
\newblock In \emph{Computer Vision and Pattern Recognition (cs.CV)}, 2018.
\newblock URL \url{arXiv:1812.03282}.

\bibitem[Yu et~al.(2020)Yu, Jiang, Gong, Zhao, Guo, Zheng, Zheng, and
  Sun]{conditionalEmbedding}
Fufu Yu, Xinyang Jiang, Yifei Gong, Shizhen Zhao, Xiaowei Guo, Wei-Shi Zheng,
  Feng Zheng, and Xing Sun.
\newblock Devil's in the details: Aligning visual clues for conditional
  embedding in person re-identification, 2020.

\bibitem[Zhang et~al.(2017)Zhang, Zhou, Lin, and Sun]{shufflenet}
Xiangyu Zhang, Xinyu Zhou, Mengxiao Lin, and Jian Sun.
\newblock Shufflenet: An extremely efficient convolutional neural network for
  mobile devices, 2017.

\bibitem[Zhang et~al.(2018)Zhang, Luo, Fan, Xiang, Sun, Xiao, Jiang, Zhang, and
  Sun]{alignedreid}
Xuan Zhang, Hao Luo, Xing Fan, Weilai Xiang, Yixiao Sun, Qiqi Xiao, Wei Jiang,
  Chi Zhang, and Jian Sun.
\newblock Alignedreid: Surpassing human-level performance in person
  re-identification, 2018.

\bibitem[Zheng et~al.(2019)Zheng, Deng, Sun, Jiang, Guo, Yu, Huang, and
  Ji]{pyramid}
Feng Zheng, Cheng Deng, Xing Sun, Xinyang Jiang, Xiaowei Guo, Zongqiao Yu,
  Feiyue Huang, and Rongrong Ji.
\newblock Pyramidal person re-identification via multi-loss dynamic training.
\newblock In \emph{Proceedings of the IEEE/CVF Conference on Computer Vision
  and Pattern Recognition (CVPR)}, June 2019.

\bibitem[Zheng et~al.(2015)Zheng, Shen, Tian, Wang, Wang, and Tian]{Market1501}
Liang Zheng, Liyue Shen, Lu~Tian, Shengjin Wang, Jingdong Wang, and Qi~Tian.
\newblock Scalable person re-identification: A benchmark.
\newblock pages 1116--1124, 12 2015.
\newblock \doi{10.1109/ICCV.2015.133}.

\bibitem[Zhou et~al.(2021)Zhou, Kang, Jin, Yang, Lian, Hou, and Feng]{DeepViT}
Daquan Zhou, Bingyi Kang, Xiaojie Jin, Linjie Yang, Xiaochen Lian, Qibin Hou,
  and Jiashi Feng.
\newblock Deepvit: Towards deeper vision transformer.
\newblock \emph{CoRR}, abs/2103.11886, 2021.
\newblock URL \url{https://arxiv.org/abs/2103.11886}.

\bibitem[Zhou and Shao(2018)]{ViewpointVehicleReID}
Yi~Zhou and Ling Shao.
\newblock Viewpoint-aware attentive multi-view inference for vehicle
  re-identification.
\newblock In \emph{Computer Vision and Pattern Recognition (cs.CV)}, 2018.

\bibitem[Zhu et~al.(2019)Zhu, Jiang, Zheng, Guo, Huang, Zheng, and
  Suno]{Viewpoint}
Zhihui Zhu, Xinyang Jiang, Feng Zheng, Xiaowei Guo, Feiyue Huang, Weishi Zheng,
  and Xing Suno.
\newblock Viewpoint-aware loss with angular regularization for person
  re-identification.
\newblock In \emph{Computer Vision and Pattern Recognition (cs.CV)}, 2019.
\newblock URL \url{arXiv:1912.01300}.

\end{thebibliography}
}

\end{document}